\documentclass[sn-basic]{sn-jnl}

\usepackage{graphicx}%
\usepackage{multirow}%
\usepackage{amsmath,amssymb,amsfonts}%
\usepackage{amsthm}%
\usepackage{mathrsfs}%
\usepackage[title]{appendix}%
\usepackage{xcolor}%
\usepackage{textcomp}%
\usepackage{manyfoot}%
\usepackage{booktabs}%
\usepackage{algorithm}%
\usepackage{algorithmicx}%
\usepackage{algpseudocode}%
\usepackage{listings}%
\usepackage{titlesec}

\theoremstyle{thmstyleone}%
\theoremstyle{thmstyletwo}%

\theoremstyle{thmstylethree}%

\begin{document}

\title[Article Title]{Recent advancements in computational morphology : A comprehensive survey}


\author*[1]{\fnm{Jatayu} \sur{Baxi}}\email{jatayubaxi.ce@ddu.ac.in}
\equalcont{These authors contributed equally to this work.}

\author[2]{\fnm{Brijesh} \sur{Bhatt}}\email{brij.ce@ddu.ac.in}
\equalcont{These authors contributed equally to this work.}

\affil*[1,2]{\orgdiv{Department of Computer Engineering}, \orgname{Dharmsinh Desai University}, \orgaddress{\street{College Road}, \city{Nadiad}, \postcode{387001}, \state{Gujarat}, \country{India}}}


\abstract{Computational morphology handles the language processing at the word level. It is one of the foundational
tasks in the NLP pipeline for the development of higher level NLP applications. It mainly deals with the processing of words and word forms. Computational Morphology addresses various sub problems such as morpheme boundary detection, lemmatization, morphological feature tagging, morphological reinflection etc. In this paper, we present exhaustive survey of the methods for developing computational morphology related tools. We survey the literature in the chronological order starting from the conventional methods till the recent evolution of deep neural network based approaches. We also review the existing datasets available for this task across the languages. We discuss about the effectiveness of neural model compared with the traditional models and present some unique challenges associated with building the computational morphology tools. We conclude by discussing some recent and open research issues in this field.}




\maketitle

\section{Introduction}\label{sec1}
The morphology of a language deals with the words. The language can be processed computationally at the word level if significant information about the words is available. This information includes structure of the word and rules to formulate a proper word. The focus of the morphology study is to understand the internal structure of the words, including the meaning associated with constituent parts and study about the combination of different units in order to make a valid word. The primary purpose of computational morphology is systematic study of language morphology using various computational methods. 

We first understand the difference between various terms related to language morphology. A stem is  the form of a word before any inflectional affixes are added. The stem may or may not be a valid dictionary word. The Root or Lemma represents base or dictionary form of a word. It is always a valid word. Depending upon the complexity of the language morphology, these terms may be used interchangeably or can be defined with clearer distinction. For example, in English, the concept of root and stem is similar since for most of the words, root and stem are the same but in arabic, most of the words have three letter root but they have two stems ; past and present. Table \ref{tab:langanalysis} shows the examples of how stem,root and lemma can defer in English, Arabic, Hindi and Gujarati language.
\begin{table}[]
\begin{tabular}{|p{0.1\linewidth}|p{0.15\linewidth}|p{0.1\linewidth}|p{0.1\linewidth}|p{0.1\linewidth}|p{0.15\linewidth}|}
\hline
Language & Word               & Stem   & Root   & Lemma  & Remarks                                    \\ \hline
English  & Running            & Run    & Run    & Run    & Stem, Root and Lemma   are same.           \\ \hline
Arabic   & daras (To Study)   & d-r-s  & daras  & daras  & Root and Lemma are   same.                 \\ \hline
Hindi    & likhana( To Write) & likh   & likhna & likhna & Stem and Root are   same.                  \\ \hline
Gujarati & Chokaro(Boy)       & Chokar & Chokro & Chokru & Stem, Root and Lemma,   all are different. \\ \hline
\end{tabular}
\caption{\textbf{Difference between various morphological terms in different languages}}
\label{tab:langanalysis}
\end{table}

Computational morphology mainly consist of tasks at two different levels: analysis and generation \cite{https://doi.org/10.48550/arxiv.2105.09404}. For the analysis, the goal is to learn about the word structure. Analysis level tasks are morphological tagging and morphological segmentation. For the generation level tasks, the focus is on generating a correct word form for example, morphological reinflection. In essence, below are some of the key tasks related to computational morphology :
\begin{itemize}
    \item Morphological segmentation: Given an inflected word, the task is to separate the root and the inflection part. For example, the segmentation of the word doing is do + ing. There can be ambiguities possible during the segmentation, for example, some words are properly separable (playing $->$ play + ing ) but some are not properly separable ( swam $->$ swim + ed) and some are separable in different ways ( Sitting $->$ Sit + t + ing, Sitting $->$ Sitt+ ing, Sitting $->$ Sit + ting ).\cite{cotterell-etal-2016-morphological-segmentation}\cite{ruokolainen-etal-2016-comparative}
    \item Lemmatization: Lemmatization is the task of transforming an inflected word to the normalized form, which is usually the dictionary or the base form of the word \cite{Plisson2004ARB, xusainova2023methods}.
    \item Morphological tagging : Morphological tagging refers to the process of assigning a morphosyntactic description (MSD) to a word \cite{cotterell-heigold-2017-cross}. Typically, morphosyntactic description is an extension of the part of speech tag. It includes POS tag along with the features like gender, number, case, tense etc. For example, the morph tag for the word runs is V;PRES;3P;SG.( Verb, Present tense, third person, singular )
    \item Morphological (re)inflection : When the root word along with the MSD tag is given, the morphological inflection task generates the required word form \cite{kodner-etal-2023-morphological}. For example, walk + V;3;SG;PRS = walks. In some cases, the given word form may not be root form, in that case the task becomes morphological reinflection \cite{cotterell-etal-2016-sigmorphon, batsuren-etal-2022-sigmorphon, goldman-etal-2023-sigmorphon}. For example, going + V;3;SG;PRS = go. Lemmatization can be treated as special case of morphological reinflection in which the target word form is always lemma ( root ) of the word.
    \item Other tasks : Some other tasks related to computational morphology have also been explored by the researchers. One such task is morphological disambiguation in which the task is to identify correct analysis when a word has different morphological analyses \cite{https://doi.org/10.48550/arxiv.1806.03740}. The historical text normalization uses morphology knowledge to convert historical word form into their equivalent modern forms \cite{bollmann-sogaard-2016-improving} \cite{https://doi.org/10.48550/arxiv.1804.02545}. Some specific tasks related to derivational morphology are explored in \cite{vylomova-etal-2017-context}\cite{deutsch-etal-2018-distributional}\cite{https://doi.org/10.48550/arxiv.1708.09151}.
\end{itemize}
\begin{figure}
    \centering
    \includegraphics[scale=0.55]{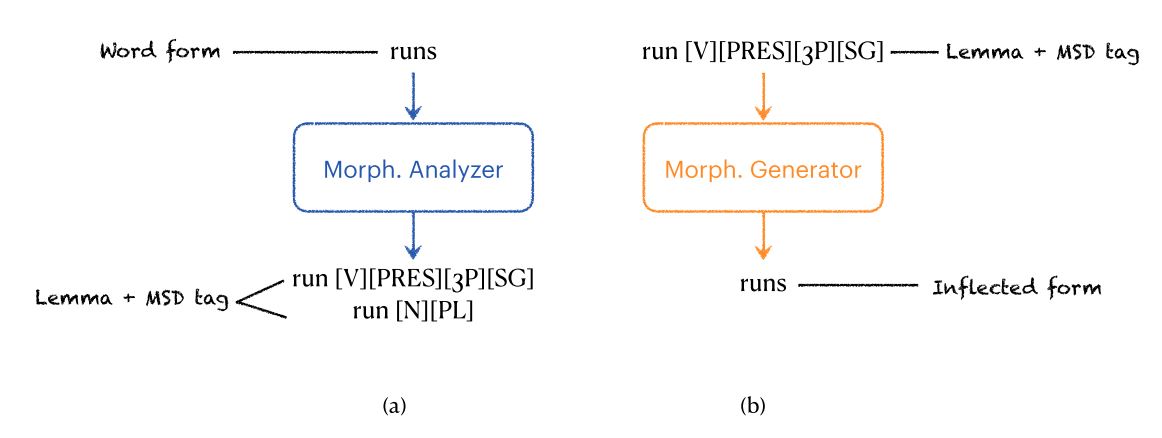}
    \caption{(a) Morphological analyzer which produces lemma and MSD tags for the given inflected word. (b) Morph generator produces inflected form when the lemma and MSD tags are given as input ( Source : \cite{https://doi.org/10.48550/arxiv.2105.09404})}
    \label{fig:MA_example}
\end{figure}

Usually, the task of morph tagging and lemmetization is jointly refereed as morphological analysis. Figure \ref{fig:MA_example} shows the example of morphological analysis and generation.

From the NLP point of view, computational morphology plays a key role as a data preprocessing step for the downstream tasks like language modelling \cite{10.1162/tacl_a_00365}, information retrival, machine translation \cite{tamchyna-etal-2017-modeling}, dependency parsing \cite{10.1162/tacl_a_00144} etc. Especially for the morphologically rich languages, high quality morphological analyzer is essential and very helpful \cite{vania-etal-2018-character} \cite{klein-tsarfaty-2020-getting}. In the current times where the NLP application pipeline is driven by pre-trained language models, it has been observed that for the embedding generation and tokenization purpose, morphological segmentation performs better than the statistical subword tokenization for morphologically rich languages \cite{nzeyimana-niyongabo-rubungo-2022-kinyabert}. The efforts for language documentation for endangered languages can also be benefited by the morphological analyzer \cite{moeller-etal-2020-igt2p, aronoff2022morphology}.

The study of computational morphology has been important task from the linguistic perspective for ages. From the early stages of the language study, various approaches and models have been evolved as highlighted in Table \ref{chronology}. In \cite{Dash2021}, authors observe that currently it is hard to conclude whether a system for complete morphological processing of a language is ready (a) which gives hundred percentage accurate results, (b) Which can be used for any downstream task, and (c) which is language agnostic. More research needs to be done in this area for creating robust computational morphology tools.

As highlighted in Table \ref{chronology} above, computational morphology consists of multiple tasks. Our focus in this paper is on the survey of various methods for the morphological analysis task. For the better understanding of other tasks, we refer the readers to \cite{https://doi.org/10.48550/arxiv.2105.09404}.

The paper is organized as follows. In section 2, we survey about morphological tools and their NLP applications across various languages. In rest of the paper, we survey about various methods for the developement of morphological tools. We survey about the rule based techniques in section 3, In section 4, we focus on the work done through machine learning and statistical methods. We demonstrate the recent advances through deep learning architectures for computational morphology in section 5. In section 6, we explore various datasets available for morphological analysis task. In section 7, we emphasis on comparative analysis of traditional vs deep learning based methods. In section 8, we present some open research issues and throw some light on possible future research directions in the field of computational morphology.
\begin{table}[h]
\centering
\begin{tabular}{|p{0.2\linewidth}|p{0.3\linewidth}|p{0.3\linewidth}|}
\hline
\multicolumn{1}{|c|}{Time   Period} & \multicolumn{1}{c|}{Dominating   Approaches}                                  & Notable Work                                                                                                                                                                               \\ \hline
1980-1990                             & Two level morphology,   Stemmer based approaches                              & {\cite{koskenniemi1983two}},   {\cite{lun1983two}}, {\cite{article_1}}, {\cite{porter1980algorithm}}                                                                                       \\ \hline
1990-2000                           & Two level morphology,   Finite state transducer                               & {\cite{oflazer1994two}},   {\cite{beesley1998arabic}}, {\cite{koskenniemi1996finite}}                                                                                                      \\ \hline
2000-2010                           & Suffix Stripping   approach, Paradigm based approach, Unsupervised morphology & {\cite{eryiugit2004affix}},   {\cite{bapat2010paradigm}},{\cite{jena2011developing}},{\cite{goldsmith2001unsupervised}},   {\cite{hammarstrom2011unsupervised}}                            \\ \hline
2010-2018                           & Supervised machine   learning, Statistical methods                            & {\cite{kumar2009morphological}},{\cite{abeera2010morphological}},   {\cite{malladi2013statistical}}, {\cite{mokanarangan2016tamil}}                                                        \\ \hline
2018 onward                         & Deep learning based   methods                                                 & {\cite{cotterell-heigold-2017-cross}},   {\cite{malaviya-etal-2018-neural}}, {\cite{akyurek-etal-2019-morphological}},   {\cite{baxi-bhatt-2021-morpheme}}, {\cite{kondratyuk-2019-cross}} \\ \hline
\end{tabular}
\caption{Overview of the approaches used during different time periods for developing computational morphology tools for various languages}
\label{chronology}
\end{table}
\section{Morphological Tools Overview and Applications in NLP}
Morphological tools plays pivotal roles in the creating of many NLP applications. Worldwide, many such tools are developed and made publicly available.Table \ref{tab:popularmorphtools} presents an overview of various popular morphological tools utilized in computational morphology. The list includes a range of analyzers, from language-specific to those with multilingual capabilities, providing a glimpse into the diversity of approaches and linguistic coverage. Tools related to the computational morphology such as morphological analyzer plays crucial role in the development of NLP tools.
\begin{table}[]
\begin{tabular}{|p{0.3\linewidth}|p{0.3\linewidth}|p{0.3\linewidth}|}
\hline
Morphological   Tool                                               & Approach/Specification          & Target   Language(s)                                   \\ \hline
Xfst   \cite{ranta-1998-multilingual}                              & Finite State   Transducer based & English                                                \\ \hline
Hindi   Morphological Analyzer and Generator \cite{goyal2008hindi} & Rule based                      & Hindi                                                  \\ \hline
Foma   \cite{hulden-2009-foma}                                     & Finite State   Transducer based & English                                                \\ \hline
Trmorph   \cite{LTEKIN10.109}                                      & Finite State   Transducer based & Turkish                                                \\ \hline
MarMoT   \cite{mueller-etal-2013-efficient}                        & Conditional Random   Field      & Czech, English,   German, Hungarian, Latin, Spanish    \\ \hline
MorphoDiTa   \cite{strakova14}                                     & Rule based                      & Czech, English                                         \\ \hline
Stanford NLP   \cite{monroe2014word}                               & Rule based                      & English, Arabic,   Chinease                            \\ \hline
Morfessor   \cite{smit-etal-2014-morfessor}                        & Probabilistic methods           & English, Estonian,   Finnish, German, Swedish, Turkish \\ \hline
SMA++   \cite{srirampur2014statistical}                            & Statistical                     & Hindi, Urdu , Telugu,   Tamil                          \\ \hline
IndiLem   \cite{chakrabarty2014indilem}                            & Unsupervised                    & Bengali, Hindi                                         \\ \hline
Apertium   \cite{forcada-tyers-2016-apertium}                      & Finite State   Transducer based & Turkic languages                                       \\ \hline
UDPipe   \cite{straka-etal-2016-udpipe}                            & Dynamic Programming   algorithm & All UD treebank   languages                            \\ \hline
CALIMA-Star   \cite{taji-etal-2018-arabic}                         & Rule based                      & Arabic                                                 \\ \hline
Mlmorph   \cite{thottingal-2019-finite}                            & Finite State   Transducer based & Malayalam                                              \\ \hline
MT-DMA   \cite{jha2019multi}                                       & Deep learning                   & Hindi, Urdu                                            \\ \hline
Ibn-Gini   \cite{10.1145/3639050}                                  & Rule based                      & Arabic                                                 \\ \hline
Morfologik                                                         & Finite State   Transducer based & Polish                                                 \\ \hline
Kuromoji                                                           & Statistical                     & Japanese                                               \\ \hline
\end{tabular}
\caption{\textbf{Morphological Analyzer Landscape: Tools and Supported Languages}}
\label{tab:popularmorphtools}
\end{table}
\textbf{Below are some popular applications of computational morphology tools in the NLP domain :}
\begin{itemize}
    \item Breaking down inflected word into root form helps in the syntactic processing and entity recognition.
    \item Information retrieval : Search accuracy can be improved by considering different word forms of a given word.
    \item POS tagging : Morphological features help in identifying POS category of a word. In this way, accuracy of existing POS taggers can be improved.
    \item Spell checking and corrections : Improving spell check systems by considering word forms and suggesting corrections based on morphological patterns.
    \item Text Generation : While automatically generating text, morphological tools can help in generating proper word forms and inflections.
    \item Better language understanding and parsing.
\end{itemize}

\section{Rule based methods}
Rule based methods have been applied for computational morphology since the earliest times. If the word formation rules are well defined and unambiguous, then the rule based methods perform reasonably well. These methods are also very popular as linguists can correlate very closely. In this section, we review some popular rule based methods for morphological analysis. 
\subsection{Two-Level Morphology and FST Based Approach}
Two level morphology was first introduced by \cite{ koskenniemi1983two}. In this method, phonological features are given more importance in the process of morphological analysis. The system considers two levels : surface and lexical levels. Direct letter to letter correspondence between the surface and lexical forms is taken into account \cite{ Koskenniemi1984}. The concept two level morphology is formalized based on the below observations:
\begin{itemize}
    \item Rules are defined as symbol-to-symbol constraints. In contrast with the rewrite rules, which are applied in sequential manner, these rules are applied in parallel.
    \item Either lexical or semantic or sometimes both contexts are referred at the same time.
    \item Morphological analysis and lexical lookup is performed in parallel at the same time.
\end{itemize}

In this method, a regular expression of a language is used to accept or reject the word. This system is based on the concept of finite state automata, which goes through the morphological components of the word. Mapping is done for the each letter corresponding to the surface level with one or more symbols at the lexical level. While the word analysis process is going on, consideration of phonological rules is done by giving much importance to the character conversion or changes in phonology, In this way, the link between surface and lexical levels are determined. This method was further modified in the work of \cite{ inproceedings_1988}, \cite{ lun1983two} and \cite{ article_1}.

The original two-level morphology model of finite state theory was modified to more advance and reliable implementations and were applied to numerous languages like ancient Akkadian \cite{ kataja1988} , Syriac \cite{ kiraz2000multitiered}, Hebrew and   Arabic \cite{ lavie1988applicabtlity}. Two level morphology model which was very popular in early 90s was replaced by the finite state transducer models (FST).

\begin{figure}
    \centering
    \includegraphics[scale=0.6]{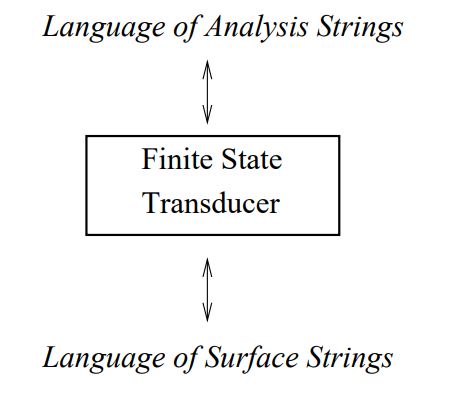}
    \caption{A morph analyzer-generator can be implemented by means of finite-state transducer. As described in Xerox convention, the analysis of the lower side language is done which is of surface string and it is represented by upper-side language. This transducer represents a data structure and the working of this structure in either direction is language-independent.(Source \cite{beesley1998arabic})}
    \label{fig:fst}
\end{figure}

FST are the structures that encode regular relations, which are mapping in the languages. It can be visualized having two sides of languages such that there is a mapping from string of one language with the string of other language. The convention indicates that the lexical transducer description compiles the analysis string of FST. This description contains morphemes and tags such as Verb, Conj, Masc, Plural etc which identify the morpheme features. These tags can be internally manipulated. The corresponding lower side language contains the surface string. Additional rules to perform deletion, epenthesis, assimilation and metathesis are encoded in the twolc ( two level compiler) language \cite{Beesley1992}. Using the finite state composition algorithm, the rules and the grammar is compiled into finite state transducers. This system can function as morphological analyzer which maps between surface string and analysis string. Figure \ref{fig:fst} shows the structure of the morphological anlyzer and generator using XFST.

Other popular methods for the implementation of FST based models are Stuttgart Finite State Transducer Tools(SFST) \cite{schmid2005programming}, Helsinki Finite State Toolkit(HFST) \cite{linden2009hfst} and Lttoolbox \cite{forcada2011apertium}.

\subsection{Paradigm based approach}
\begin{figure}
    \centering
    \includegraphics[scale=0.75]{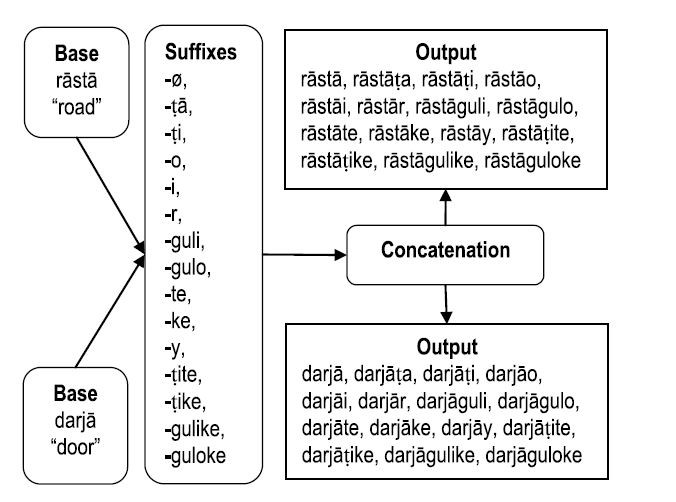}
    \caption{Information processing in paradigm based approach ( Source : \cite{Dash2021} )}
    \label{fig:paradigm}
\end{figure}

Paradigm based approach is more sophisticated way to build rule based morphological analyzers. In this method, the task is to map an input word to the corresponding paradigm. Different paradigms are constructed for various POS categories depending upon the morphological features. For a given stem and specified grammatical feature set, a paradigm defines all possible word forms that can be generated \cite{baxi-etal-2015-morphological, bhatt2011indowordnet}. For creating a paradigm, large number of inflected forms are collected and list of possible suffixes along with the morphological features are identified. The word forms which accepts similar set of suffixes are mapped into a single paradigm. Once all the paradigms are defined, any unknown inflected word can be classified into corresponding paradigm so that the grammatical features and root word formation rule can be easily identified. As shown in the Figure \ref{fig:paradigm}, the bengali words Rāstā (Road) and Darajā (Door) belongs to the same paradigm. The second half of the figure shows the possible suffixes which are common to both the words and the last part shows various inflected words that can be formed using the given suffixes.

When the inflected word is given to the system, some rules based on the suffixes are checked to determine the paradigm of the word. It may be possible that the same rule may be present in more than one paradigm and the input word may be mapped with multiple paradigms. In such situations heuristics or statistical methods may help to resolve the ambiguity.

For many Indian languages, paradigm based approach has been widely used for creating morphological analyzer  \cite{Bapat2010}\cite{baxi-etal-2015-morphological}\cite{goyal2008hindi}\cite{jena2011developing}. For in depth explanation about the paradigm based approach in context with the Indian langages we redirect the readers to \cite{gillon1995natural}.

\subsection{Stemmer based approach}
This method employs a stemmer that has a list of stem and replacement rules as well as a set of lexical rules. In this approach, we must accurately specify each of the language-specific rules. The Porter algorithm \cite{porter1980algorithm} is a well-known technique used for word stemming and to perform morphological analyses in English and many other languages worldwide. In \cite{rajeev2007morph}, authors use suffix stripping method which is a stemmer based approach to create morphological analyzer for the tamil language. The authors observe that since Malayalam language requires many morphophonemic changes during the word formation process so, the suffix stripping method may be used. Suffix stripping algorithms do not require a lookup table; they require only the rules to find root/stem.

This algorithm is effective in stemming the words. It is not much useful for lemmatization and morphological analysis. The basic requirement of morph analyzer is generating root word along with the grammatical features. It may also be the case that the stem obtained through this stemming algorithms may not be proper root word. The basic purpose of a morphological analyzer is to highlight grammatical contribution of individual morpheme in the word formation process. Thus, a stemmer may not be a good option for the morphological analysis.

\section{Machine learning based systems}
The rule based systems were initially popular for building computational morphology tools but with time, researchers realized that the rule based method had some major limitations. First of all, the linguistic rules must be well documented. For many low resource languages, the documentation containing such rules is difficult to obtain. Also, ambiguities in the word formation rules present unique challenges. The methods introduced for the disambiguation are not very effective. With the above limitations and rise of machine learning based methods, the researchers shifted their focus from traditional rule based methods to machine learning based methods. The basic idea is to formulate the morphological analysis as a machine learning problem and then to apply various supervised and unsupervised models for this task. The trained model then can be used to predict morphological features for the unknown inflected word. In this section, we survey about the various supervised and unsupervised models in the literature for the computational morphology related tasks.
\subsection{Unsupervised Models}
Unsupervised learning was very popular approach for computational morphology before the evolution of the neural models \cite{goldsmith2017computational}\cite{hammarstrom2011unsupervised}. Unsupervised approaches assumes that there is raw natural language data and the model has to learn morphological description without any supervision. Work carried out by Goldsmith \cite{goldsmith} in the field of unsupervised morphology created breakthrough and motivated many researchers to work in this field. He combined methods from various earlier works and aligned them against the background of information theory and linguistics. His focus was on unsupervised learning of inflectional morphology. In \cite{10.1162/089120101750300490}, authors note that the unsupervised approaches address the computational morphology in the following different ways 
\begin{itemize}
    \item Border and frequency : In these methods, the segmentation border is identified if a substring occurs with a variety of substrings immediately adjacent to it. Frequent substrings are also considered as candidates for the segmentation. For the given segmentation, stem-suffix co-occurrence statistics are gathered.
    \item Group and abstract :In this set of methods, grouping of morphologically related words is done based on some metric such as string edit distance. It may also include semantic features \cite{ schone2001toward} , frequency signatures \cite{ wicentowski2004multilingual} or distributional similarity \cite{ freitag2005morphology}. After that, some morphological pattern is abstracted that gets repeated among the groups. These patterns provide clues for the segmentation and can help in formation of morphological paradigms. 
    \item Features and classes : In these methods, a word is considered to be made up of a set of features such as n-gram \cite{ mayfield2003single} and Initial/terminal/mid-substring \cite{ de2007bootstrapping} Based on the occurrence of the features, a specific word or stem can be identified. To formalize this intuition, concept of TF-IDF and entropy has been used. When any unknown word is given as an input, features are used to select which word it is morphologically related to. 
    \item Intercalated morphology: In this family of methods, classification of phonemes is done into vowels and consonants. Various methods based on frequency are applied after each word is converted into vowel and consonant skeleton. These methods are based on the special kind of non-concatenative morphology which is known as intercalated morphology \cite{ xanthos2008apprentissage} \cite{ bati2002automatic} \cite{todd-etal-2022-unsupervised}.
\end{itemize}

Goldsmith \cite{goldsmith} proposed Linguistica: An automatic morphological analyzer. It has widespread application because of its language independency. The author proposes to divide the process of morphological analysis into a set of heuristics and Minimum Description Length (MDL) evaluation process. The heuristics then is divided into initial bootstrapping heuristics which determines the first analysis of stems and suffixes, and incremental heuristics which modifies this analysis. The MDL then decides whether the modifications made
by the incremental heuristics should be adopted or dropped. The bootstrap heuristics generates a set of candidate suffixes and stems for further analysis by breaking word at points of high successor frequency. The minimum description length is the quantitative measure which helps us determine which of the analysis is better.

An important milestone in the field of unsupervised morphology was the development of Morfessor \cite{morphessor10} – It provides set of methods for the segmentation of words using unsupervised approach. The initial version of Morfessor which is popular as morfessor baseline was proposed by  \cite{creutz2002unsupervised}. Its software implementation Morfessor 1.0 was developed by the authors in 2005. The cost function of Morfessor Baseline is derived using maximum a posteriori estimation. A number of Morfessor variants have been developed later, including Morfessor Categories-MAP and Allomorfessor. Although, large improvement was seen on morfessor baseline model by these algorithms in some aspects, still the baseline version is considered as stable version and widely used for the morphological analysis purpose.

In \cite{Ak2012}, authors proposed Trie based Unsupervised Morphological analyzer for Turkish and English Language. The algorithm requires corpus along with word list and word occurrence information. In this algorithm, trie is constructed that consists of the occurrences of the words and characters as nodes. After the trie has been constructed, roots of the given words are detected by scanning the occurrences in the path of the word. When the root is disclosed, new trie is created from the affix part. The recursive execution of the algorithm continues till there is no affix left to process. In \cite{yoshinaga-2023-back}, authors have developed japanese morphological analyzer using trie based approach.

In \cite{Narasimhan2015}, the authors proposed unsupervised method for uncovering Morphological chains. Log linear model is used which enables to incorporate semantic level as well as orthographic level features of the word. The meaning of Morphological chain in this context is chain of all possible word forms from the base word. For example, Play ${\rightarrow}$ Playful${\rightarrow}$ Playfully. The log linear model contains the feature $\varnothing : w \times z$ where w is word and z is candidate pair. The chain starting with the base form is constructed and likehood for the set of candidates is controlled by features such as semantic similarity, affix correlation, presence in the word list etc. The system was tested for English, Turkish and Arabic languages was performing better than the baseline morfessor system.

Unsupervised models might have performed well for some languages but in general, without the labeled data, unsupervised approaches may not be able to identify morphological structures with the same level of accuracy as supervised methods. Also, unsupervised methods may not perform well when dealing with words that are not in the training data, leading to errors in analysis.

\subsection{Supervised Models}
In the supervised models, it is assumed that some labelled corpus is available. The model for the morphological analysis primarily requires a dataset in which each word is mapped with the corresponding base(dictionary) form along with the grammatical features specified by a particular inflection. In order to train the model effectively manual feature engineering also needs to be done. The decision on which features to consider can be crucial and often language dependent. Once the feature engineering is done, different classifiers such as support vector machine (SVM), Neural network may be explored to train the system.

In \cite{5329355}, authors created Morphological analyzer for agglutinative languages using machine learning approaches. They formulated the problem as sequence labeling task. The main aim of the sequence labeling approach is to predict output y from the given x. In the training data, the input sequence ‘x’ is mapped with output sequence ‘y’. Given the task, x indicated sequence of characters and y indicates the corresponding root component or the set of grammatical features. In this work, the authors use support vector machine and conditional random field based techniques for machine learning and compare the obtained results.

\begin{figure}
    \centering
    \includegraphics[scale=0.75]{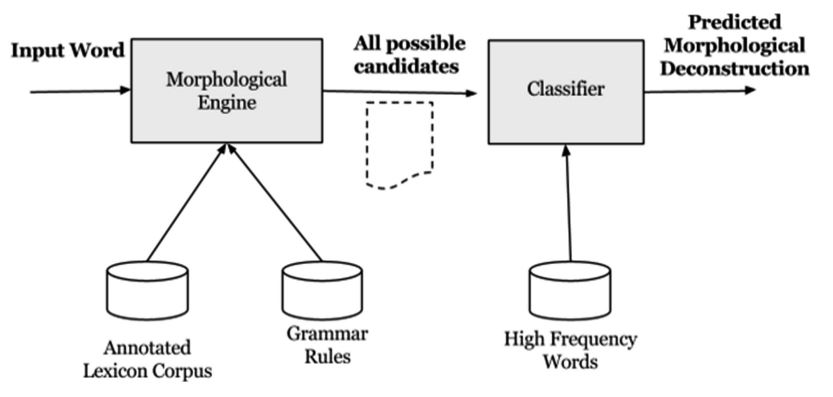}
    \caption{Supervised morphological analyzer ( source : \cite{mokanarangan2016tamil})}
    \label{fig:tamilmorph}
\end{figure}

In \cite{mokanarangan2016tamil}, the authors created morphological analyzer for Tamil language using SVM. In this approach, lexical labels are generated by breaking down the text. For the constituent elements, lemma ID and POS tags are represented. Features are identified based on the document level context. The support vector machine classifier is used to perform the classification based on the identified features. As shown in the figure \ref{fig:tamilmorph}, the input word is given to the morphological engine and using lexical rules, all possible forms are generated which are given to the SVM classifier to get the final morphological deconstruction.

Manual feature engineering and need of annotated data and rules remained major challenges in using the supervised machine learning methods. From the year 2015 onward, the evolution of neural network provided new directions for the research in this field. Researchers experimented with the deep learning models such as Recurrent Neural Network(RNN) and Long Short Term Memory(LSTM) which does not require any manual feature engineering. For many languages across the world, the neural models have outperformed the standard rule based and statistical morph analyzers. In the following section, we survey various neural architectures and dataset requirements for the computational morphology related tasks.
\section{Deep learning based approaches}
Standard rule based, statistical and unsupervised models explored in previous sections had many limitations with respect to the development of computational morphology tools. The evolution of neural network models especially deep neural network models after 2015 motivated the researchers to apply these models for the morphological analysis and generation tasks. Even though the deep neural network models are data hungry but they do not require any manual feature engineering. A task of computational morphology first needs to be converted into neural network compatible problem. The crucial part in this process is to choose a proper deep neural network architecture. Many state of the art models such as Convolution Neural Network(CNN), Recurrent Neural Network(RNN) are available \cite{bjerva2017one}. The combination of different neural network architectures depend on the type and nature of the problem and the representation of data.
\begin{figure}
    \centering
    \includegraphics[scale=0.7]{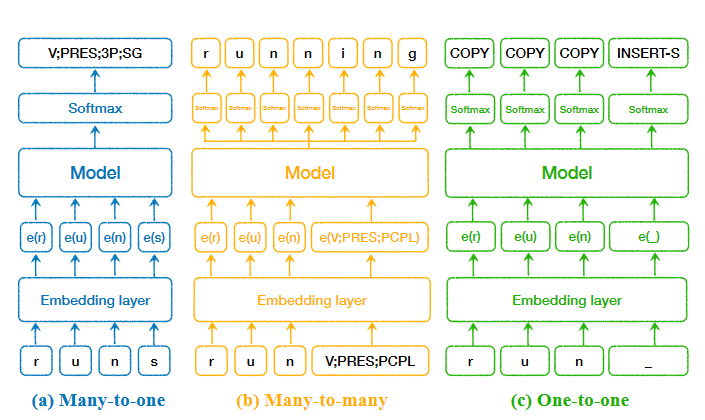}
    \caption{Different neural network learning scenarios ( Source : \cite{https://doi.org/10.48550/arxiv.2105.09404})}
    \label{fig:nn_arch}
\end{figure}

Figure \ref{fig:nn_arch} represents different neural network learning scenarios with respect to the morphological anaysis problem. The first part of the figure \ref{fig:nn_arch} shows many-to-one scenario. In this architecture, multiple input sequences are processed by the network and a single output is produced. This architecture can be useful when we have a sequence of inputs that we want to use to predict a single output. As shown in the figure \ref{fig:nn_arch}, for the morphological tagging task, the first step is to get the representation of the text in form of embeddings. The output is predicted as only one symbol or one chunk of symbol like Morpho-Syntactic Description(MSD). It is like a classification scenario. Morphological tagging can be considered as a classification task if we treat each unique MSD tag as a class.

The second part of Figure \ref{fig:nn_arch}  shows many-to-many architecture. In this architecture, multiple input sequences are processed by the network and multiple output sequences are produced. This architecture can be useful when we have a sequence of inputs that we want to use to predict another sequence of outputs. For the morphological generation task, the model first receives the input and turns it into the embedding representation before generating a second sequence of symbols, which may have the same or a different length from the input. Sequence to sequence transduction is comparable to this. With this situation, it is possible to deal with morphological inflection and morphological generation tasks.

The third part shows one-to-one architecture. This architecture can be useful when we have a single input that we want to use to predict a single output.In the one-to-one scenario, for each corresponding symbol in the input, the model makes a prediction. The typical task can be sequence labeling. For instance, morphological inflection task shown in the third half where the task is to generate the present participle for verb run. If the objective of the model is to predict the edit actions to transform the input into the corresponding output, the correct operations will be  Copy, Copy, Copy, and Insert-s.
\subsection{Various Architectures based on Deep Learning}
The model part in figure \ref{fig:nn_arch} represents one of the four type of neural network architectures: feedforward, convolutional neural networks (CNNs), recurrent neural networks (RNNs) and the transformer. In some approaches these models are also combined with other statistical models like HMMs(Hidden Markov Model), CRFs(Conditional Random Field), etc. For the classification task, usually the softmax layer is used. For the sequence transduction task, the encoder-decoder structure has achieved good success.
\subsubsection{Feedforward}
The feedforward neural networks with fully connected architecture is also called multi-layer-perceptron. They can be conveniently applied to classification problems as well as structured prediction problems. However, they have a drawback of requiring fixed length representation. One of the simple ways to represent a word using this approach is bag of word approach in which the vector representation of multiple symbols are added in order to get the vector representation of the larger unit that contains those symbols. This method suffers from the limitation that it often discards important semantic information from the given input \cite{goldberg2016primer}.
\subsubsection{Convolutional Neural Networks}
Convolutional neural networks combine local predictors to create a fixed representation of the structured input by selecting them from a broad structure. Each segment of a k-sized sliding window over the input is subjected to a filter, which is a non-linear function, in order to extract the crucial data and turn it into a fixed-size representation. The result is defined to be combined into a single representation by the pooling process. Convolutional layers may be used concurrently. The two most popular pooling methods are max pooling and average pooling. This approach has been found to produce better vector representation than CBOW(Continuous Bag of Word) \cite{goldberg2016primer}.
\subsubsection{Recurrent Neural Networks}
Recurrent neural networks are employed whenever the data is in the form of arbitrary length sequences to capture the structured information. Basically, there are two types of RNN: (1) Vanila RNN \cite{ elman1990finding} (2) Gated RNN which can be implemented using the Gated Recurrent Unit (GRU) \cite{ cho2014learning} or the Long-Short Term Memory (LSTM) architecture \cite{ hochreiter1997long}. The state at position ``i'' is represented by the vanilla RNN as a linear combination of the input at this location and the prior state that has undergone a non-linear activation (usually ReLU or tanh). Although it is efficient to record sequential information, it is challenging to train efficiently because of the vanishing gradient problem \cite{ hochreiter2001gradient}. To address the problems with a vanilla RNN, GRU and LSTM incorporate memory cells that are controlled by gating elements. The past and future context are both captured by the bidirectional RNNs' encoding of the sequence in both forward and backward orientations \cite{ schuster1997bidirectional}.
\subsubsection{Encoder-Decoder Structure and the Transformer}
The sequence-to-sequence architecture or the encoder-decoder architecture \cite{ sutskever2014sequence} was primarily designed for the machine translation but was later applied to other tasks as well. It breaks down the entire process into encoder and decoder transformations. Encoder and decoder internal architectures resemble the aforementioned systems. The encoder transforms the input sequence into a vector representation in the standard seq2seq architecture. To create the output sequence, the decoder receives the same vector representation as input at each time step, typically together with other data relevant to the task. The attention mechanism improves this concept even further. Every weight vector used for each decoding state is learned via the attention mechanism \cite{bahdanau2014neural}. 

The encoder-decoder architecture based on RNN with attention has been applied to many string transduction problems.  However, it has been observed that such models are slow for the large scale data and they also lack parallel computation. The transformer model \cite{vaswani2017attention} attempts to improve the seq2seq model to overcome its dependency on the sequential processing. Transformer does not employ the RNN architecture and is stateless. The input and output representations are computerised using self-attention layers and multi-head attention. Other recent notable contributions using transformer architecture are discussed in \cite{zundi-avaajargal-2022-word} and \cite{liu-hulden-2022-detecting}.
\begin{figure}
    \centering
    \includegraphics[scale=0.6]{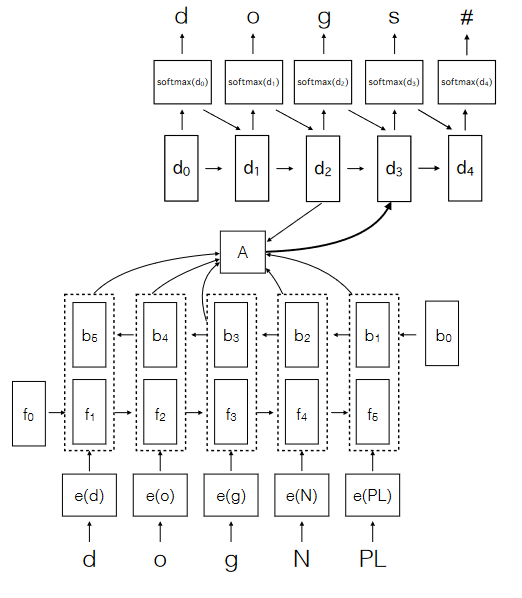}
    \caption{Illustration of the attention mechanism in the seq2seq architecture.
The bottom portion of this illustration shows a bidirectional encoder.
The focus for creating the character's attention is depicted in the figure. (Source \cite{silfverberg2017data})}
    \label{fig:my_label}
\end{figure}
\subsection{Deep Learning based models for Morphological Segmentation}
\begin{figure}
    \centering
    \includegraphics[scale=0.7]{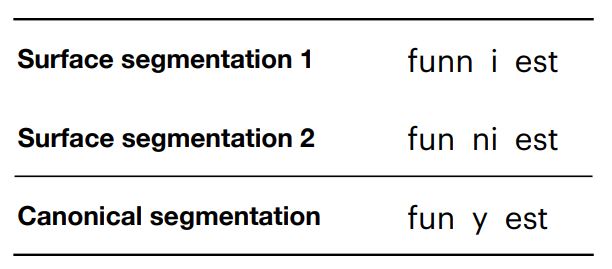}
    \caption{The canonical and surface segmentation of English word funniest : two different approaches. This example was created with reference to the example in \cite{cotterell2016joint}}
    \label{fig:surcan}
\end{figure}

In this section, we survey about the existing work done for the morphological segmentation and lemmatisation tasks using the deep neural network approaches. The advantage of deep learning based models such as RNN and LSTM is that they do not require feature engineering and can capture the language specific characteristics from the huge amount of training data. The challenging part is data representation and defining the nature of the problem such that it fits in the deep learning model framework. Usually, the segmentation task should involve derivation and composition forms along with segmenting inflection parts. For example, in \cite{cotterell2016joint}, authors do not segment the word 'touching' as 'touch + ing' instead it is treated as a single morpheme with adjective POS category. Authors of \cite{cotterell2016joint} distinguish surface segmentation and canonical segmentation. For the surface segmentation, the concatenation of the segments are exactly the original word. In canonical segmentation it may not be the case. Ambiguities arising in surface segmentation can be avoided in canonical segmentation. Figure \ref{fig:surcan} depicts two possible methods for the surface segmentation for the English word 'funniest' along with the canonical segmentation of the word.

In \cite{wang2016morphological}, authors have described the surface morphological segmentation. They treat the task as sequence labelling problem. Each character is classified as beginning, middle, end or a single character morpheme. The authors proposed three LSTM architectures : Simple window, a multi window and bidirectional multiwindow LSTM model. The experiments were done on Hebrew and Arabic data. The bidirectional multi-window LSTM model performs significantly better than the standard LSTM model and is also capable of producing outcomes that are on par with non-neural algorithms. 

In \cite{kann2018fortification}, authors concentrate on the polysynthetic language morphological segmentation task with limited labelled data. By utilizing unlabeled data, they make improvement in the character-based seq2seq model with an attention mechanism. They use the concept of data augmentation method. They study the cross-lingual transfer effect in which one model is trained for all languages. In conclusion, authors were able to significantly outperform the CRF baseline with thier proposed model. 

The work of \cite{moeller2018automatic} is specifically for the low resource languages. They analyze the morphological segmentation and labeling task with respect to language documentation. They use the sequence labelling approach similar to \cite{wang2016morphological} but their labels are much more detailed. No separate symols were used for representing a single character morpheme. They utilise a multi-class SVM to categorise the morphemes to their tags and a CRF to forecast the BIO(Beginning-Inside-Outside) tags.
It has been determined through numerous studies that non-neural models with linguistic feature engineering still have an advantage in situations where there is a very small amount of labelled data \cite{kann2018fortification} \cite{wolf2018structured}.

Semantics are incorporated into segmentation in the work \cite{cotterell2018joint}.
The model consists of three components: a transduction factor, a segmentation factor, and a composition factor. For the composition factor, they conducted experiments with addition to RNN in order to combine the morpheme embeddings to roughly estimate the vector of the entire word. The transduction factor is a finite state machine which is weighted in nature and which computes probabilistic edit distance by considering an additional input and output context.
They discovered that segmentation performance is enhanced by concurrently modelling semantic coherence and segmentation. 

In \cite{cotterell2016joint}, authors described the canonical segmentation task, which breaks down a word into a series of canonical segments. To approximate the gradient for learning, it uses a sampling approach, a semi-markov segmentation factor, and a finite-state transduction factor. An external dictionary is used for validation. They tested on the languages of English, German, and Indonesian and produced a dataset. \cite{kann2016neural} and \cite{ruzsics2017neural} both discuss attention-enhanced encoder-decoder models, in which the work of morphological segmentation is viewed as a string transduction problem. In \cite{kann2016neural}, the encoder-decoder paradigm, which makes use of external dictionaries and resources, is supplemented by a neural re-ranker. They obtained cutting-edge performance when comparing the results with \cite{cotterell2016joint}. The context information is not considered in any of the work previously discussed. When a word has different possible segmentations, the context is a vital source of information for disambiguation.

Lemmatization is a process of obtaining root ( base ) form of an inflected word. This problem can be solved in one of two ways: as a string transduction problem by converting the input character string of the inflected word forms to the corresponding lemma string, or as a problem of classification by choosing an edit tree that generates the lemma from the given inflected form. With two bidirectional RNNs, the authors of \cite{chakrabarty-etal-2017-context} treat this problem as edit tree classification. The first one creates syntactic embeddings, and the second one learns to represent the local context and predicts the lemma with the highest likelihood. They tested LSTM and GRU and found that both performed admirably. Compared to Lemming \cite{muller2015joint}  and Morfette \cite{chrupala2008learning}, the system performs better.

\subsection{Deep Learning based models for Morphological Tagging}
Morphological tagging is usually treated as a sequence labeling problem, in which for each word in the sentence, the tagger is required to predict the corresponding MSD tag. It is usually one-to-one learning scenario. For accurate tagging, the context information is critical.

The neural model for this task typically has three components: word representation, context encoder and a tagger. In \cite{labeau2015non}, authors make use of convolution layer to derive the word representation and compare the character based embeddings with word embeddings. For the context encoding, they use feed forward and bidirectional LSTM architectures. For tagging, they compare two approaches: a simple softmax prediction where a simple softmax function is applied to the output at each time step  to predict corresponding tag; and a structures prediction approach in which \cite{viterbi1967error} algorithm is used. The experiments are conducted using the German TIGER Treebank corpus \cite{brants2002tiger}. Combining character and word based embeddings produced the exceptional results. Word representation on various models, such as feed forward, CNN, CNNHighway, LSTM, and bidirectional LSTM, is the emphasis of \cite{2016arXiv160606640H} and \cite{heigold-etal-2017-extensive}. They anticipate the tags using a straightforward softmax layer. The LSTM-based model was found to perform marginally better than CNN-based models.

\subsection{Joint learning of lemma and morphological tag}
Lemmatization and morph tagging can be mutually dependent and can resolve ambiguities where there are different ways for lemmatization or tagging. It is common to perform the lemmatization and morphological tagging together which is referred to as joint learning or morphological analysis. Table \ref{joint} shows the example of jointly learning lemma and MSD. \cite {muller2015joint} provides empirical evidence that the joint learning is mutually beneficial. Other popular work related to joint learning are \cite{kondratyuk2018lemmatag}, \cite{malaviya-etal-2018-neural} and \cite{ mccarthy2019sigmorphon}

LemmaTag \cite{kondratyuk2018lemmatag} is a system for jointly generating MSD tag and lemma using bidirectional RNN. It has three parts : (1) The encoder in which the word embeddings are generated by concatenating character embeddings. It is combined with word level embeddings for better representation. (2) The tagger applies fully connected layer to obtain the tags. (3) The lemmatizer module creates the lemma form character by character using an RNN decoder with soft attention applied to character encodings, summed embeddings for the current word, predicted tag features, and context representations. The weighted total of the losses experienced by the tagger and the lemmatizer is the final loss function. They discovered that learning lemmatization and morphological tagging simultaneously by sharing encoder settings and feeding the lemmatizer with the expected tags improves the efficiency of both the processes, especially for morphologically rich languages. 

The system developed by \cite{malaviya-etal-2018-neural} has similar architecture as \cite{vylomova2019contextualization}. The lemmatizer is a seq2seq model with hard attention mechanism. For lemmatization, greedy decoding and crunching scheme is experimented. It is a heuristic to approximate true joint learning by first predicting k-best tags and then incorporating it into the lemma generation process. It is observed that the joint learning is more helpful when the training data is limited to lemmatization. The accuracy is much better when the lemmatizer is supplied the ground truth MSD tags.  The proposed system outperforms other similar work in \cite{bergmanis2018context} \cite{ chrupala2008learning} \cite{muller2015joint}.

In \cite{akyurek-etal-2019-morphological}, authors present another such work. The word embeddings in this system are obtained at the character level using a unidirectional LSTM. A bidirectional LSTM is employed for the context representation. They employ an LSTM to produce the lemma characters and the MSD features. In low-resource settings and for morphologically more complicated languages, the authors note that forcing the decoder to jointly anticipate the lemma form and the factored MSDs yields generally lower results than decoding merely the MSDs.

\begin{table}[]
\begin{tabular}{|l|p{0.5\linewidth}|}
\hline
Input & He told the children an encouraging story.                                         \\ \hline
Lemma & He tell the child a encourage story.                                               \\ \hline
MSD   & [PRON;NOM;SG, V;PST, DET;DEF, N;PL,DET;IND, ADJ, N;SG, PUNCT]                      \\ \hline
      &                                                                                    \\ \hline
Input & He is encouraging the children with a story.                                       \\ \hline
Lemma & He be encourage the child with a story.                                            \\ \hline
MSD   & [PRON;NOM;SG, AUX;SG;3;PRS, V;PRS;PCTP,  DET;DEF, N;PL, ADP, DET;IND, N;SG, PUNCT] \\ \hline
\end{tabular}
\caption{\textbf{Joint learning of lemma and morphological tag examples}}
\label{joint}
\end{table}

\subsection{Transformer based approaches}
BERT (Bidirectional Encoder Representations from Transformers) is a transformer-based neural network architecture for natural language processing tasks such as language understanding and machine translation. Since its release in 2018, BERT has become a popular model architecture for many NLP tasks and has been the basis for many subsequent models. The main advantage of BERT is Pre-training. BERT is pre-trained on a large dataset of unannotated text, which means it can be fine-tuned for specific tasks with relatively small amounts of labeled data \cite{devlin2018bert}.

One way to use BERT for morphological analysis is to fine-tune a pre-trained BERT model on a morphological task, such as morphological tagging or lemmatization. For example, one can fine-tune BERT on a dataset of words and their corresponding morphological tags (such as noun, verb, adjective, etc.) or lemmas (the base form of a word). Another way to use BERT for morphological analysis is to use it as a feature extractor for other models. For example, one can use the BERT embeddings as input to a neural network model that performs morphological analysis. BERT's ability to capture context and the fine-tuning capability makes it a powerful tool for morphological analysis, as it allows the model to understand the morphological structure of words and how they are used in context.

\begin{figure}[h]
    \centering
    \includegraphics[scale=0.8]{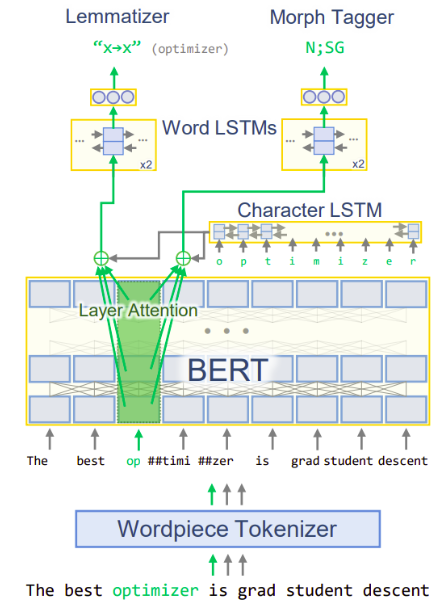}
    \label{bert}
    \caption{Morphological analyzer based on BERT. The mBERT model acts as a multilingual encoder and the word level and character level LSTMs are the decoders (Source : \cite{kondratyuk2019cross})}
\end{figure}

In \cite{kondratyuk2019cross}, authors use the pre-trained multilingual BERT cased model to encode input sentences before adding additional word- and character-level LSTM layers. Lemmas and morphological tags are then simultaneously decoded using simple sequence tagging layers. Lemmatization decoding is viewed as a sequence classification problem, and in order to find the edit actions required to convert an inflected word to its lemma, a feed-forward layer is applied to the lemmatizer LSTM final layer. In order to jointly forecast factored and unfactored MSDs for morphological tagging, a feed forward layer is used. Figure 8 shows the architecture of their system.

Another transformer based work is \cite{straka-etal-2019-udpipe}. In this work, first the input words are converted into embeddings followed by three shared bidirectional LSTM followed by softmax classifier to generate morphosyntactic features. A pre-trained contextual word embeddings(BERT) is added as the another input to the system.

The typical SeqtoSeq models may not perform effectively in low resource environments due to a lack of labeled example data. In \cite{ saunack-etal-2021-low}, authors state that the primary objective is to determine how well cross-lingual transfer works for the task of lemmatization. The authors' LSTM-based encoder-decoder model employs a two-step attention procedure. They note that a monolingual model trained on roughly 1000 training samples provides competitive accuracy for the majority of Indian languages. They also note that having POS tags present is a feature that helps with training. 

\section{Datasets for Morphological processing}
As surveyed in the above sections, the methods for creating morphological analyzer are either rule based or machine learning based. For the rule based methods, knowledge is required in the form of linguistic rules, suffix tables, and other language-specific characteristics. For the machine learning based methods focused on deep neural network architectures, they do not need any manual feature engineering but on the other hand they require massive amount of training data for the effective and accurate results \cite{hamalainen-etal-2021-neural}. It is expected that the dataset should be annotated in a standard well accepted format so that it remains consistant and generalized. Each word should be mapped with its associated root word as well as the potential collection of grammatical attributes in a dataset of this type \cite{kirov-etal-2018-unimorph}. Table \ref{tab:dbsurvey} shows various datasets used for creating morphological tools in different languages. The majority of the current research in the area of morphological analysis for a variety of languages uses datasets from either the Unimorph schema \cite{kirov-etal-2018-unimorph} or the Universal Dependencies Treebank \cite{nivre-etal-2016-universal}\cite{nivre2020universal}. 

The goal of the Universal Dependencies (UD) project is to create treebank annotation of morphology and syntax that is cross-linguistically consistent for many languages.
The dataset's initial release in 2015 included 10 treebanks spread over 10 languages.
Version 2.7, which was released in 2020, has 183 treebanks spread across 104 languages.
The annotation consists of Lemmas, dependency heads, universal dependency labels, Feats (universal morphological characteristics), universal part-of-speech tags (UPOS), and language-specific part-of-speech tags (XPOS) \cite{10.1162/coli_a_00402}.

The Universal Morphology (UniMorph) project is a collaborative attempt for providing broad-coverage of normalized morphological inflection tables for hundreds of diverse world languages. The project comprises two major parts: a language-independent feature schema for the rich morphological annotation and a type-level resource of annotated data in diverse languages realizing that schema. Over 212 features and 23 different levels of meaning are included in the UniMorph schema. Morphological categories like person, number, tense, and aspect are the dimensions of meaning. Each one illustrates an organized semantic area in inflectional morphology. You can find them in the following categories: Aktionsart, animacy, aspect, case, comparison, definiteness, deixis, evidentiality, finiteness, gender, information structure, interrogativity, mood, number, part of speech, person, polarity, politeness, switch-reference, tense, valency, and voice. There are various numbers of features in these dimensions, ranging from 2 for finiteness to 39 for case. The most precise meaning distinctions that may be made within a given dimension are represented by features. The article \cite{batsuren2022unimorph} highlights recent changes to the Unimorph schema, including the addition of new low resource languages such as Gujarati \cite{baxi-bhatt-2022-gujmorph}, a hierarchical organization inside the schema, morpheme segmentation for 16 languages, the addition of derivational morphology, etc.
\begin{table}[]
\begin{tabular}{|l|l|}
\hline
Name                                                           & Target   Language(s)               \\ \hline
Unimoprh   \cite{batsuren2022unimorph}                         & 169 languages                      \\ \hline
UD-Treebank   \cite{nivre-etal-2016-universal}                 & 148 languages                      \\ \hline
Mighty-Morph   \cite{goldman-tsarfaty-2022-morphology}         & English,Germen,   Hebrew, Turkish  \\ \hline
Neural-Morphology-Dataset   \cite{hamalainen-etal-2021-neural} & Germen, Finnish and   20 others    \\ \hline
MorphyNet   \cite{batsuren-etal-2021-morphynet}                & Russian, Hungarian   and 13 others \\ \hline
\end{tabular}
\caption{\textbf{Details of various morphological datasets}}
\label{tab:dbsurvey}
\end{table}

\section{Discussion : Traditional vs Deep learning approaches}
\begin{table}[]
\begin{tabular}
{|p{0.2\linewidth}|p{0.2\linewidth}|p{0.2\linewidth}|p{0.2\linewidth}|p{0.2\linewidth}|}
\hline
\multicolumn{1}{|c|}{\textbf{Approach}} & \multicolumn{1}{c|}{\textbf{Methodology}}                                                                                     & \multicolumn{1}{c|}{\textbf{Data   Requirements}}                                & \multicolumn{1}{c|}{\textbf{Advantages}}                                    & \multicolumn{1}{c|}{\textbf{Limitations}}                                         \\ \hline
Rule based                              & Based   on predefined linguistic rules and patterns                                                                           & Comprehensive   linguistic rules and patterns designed by experts or linguists.  & Transparent   and interpretable due to explicit rules                       & Time-consuming   and labor-intensive to create rules                              \\ \hline
Finite state transducer based           & Utilizes   finite state machines to model and recognize complex linguistic patterns                                           & Datasets   representing the finite set of rules or regularities in the language. & Efficient   for specific linguistic phenomena with well-defined rules.      & Limited   expressiveness in capturing certain linguistic intricacies.             \\ \hline
Statistical                             & Utilizes   probability and statistical models to analyze and make predictions based on   the frequency of linguistic patterns & Large   corpora to establish accurate probability models                         & Effectiveness   in capturing statistical regularities in the language.      & Difficulty   in capturing rare linguistic patterns                                \\ \hline
Supervised Machine learning             & Trains   models using labeled data                                                                                            & Requires   labeled datasets                                                      & Provides   precise predictions when trained with high-quality labeled data. & Limited   generalization to unseen patterns                                       \\ \hline
Unsupervised Machine learning           & Involves   training models on unlabeled data to identify patterns and structures                                              & Large,   diverse, and unlabeled datasets                                         & Ability   to identify hidden patterns without labeled data,                 & Difficulty   in capturing rare linguistic patterns                                \\ \hline
Deep Learning                           & Employs   neural networks with multiple layers to automatically extract features                                              & Requires   large volumes of data                                                 & Potential   for high accuracy in tasks with large amounts of data.          & Computationally   intensive, Overfitting possible when dealing with limited data. \\ \hline
\end{tabular}
\caption{\textbf{Comparative Analysis of Morphological Analyzer Approaches: Methods, Advantages, and Limitations}}
\label{tab:compare}
\end{table}
In this section, we first compare various approaches discussed so far in this paper and then discuss the advantages of deep learning approaches compared to traditional approaches. Table \ref{tab:compare} shows the comparative analysis of various approaches discussed in this paper. Traditionally, the morphological analysis was considered as a language specific problem because it deals with linguistics and language specific rules and every language have different rules for word formation. Due to this fact, approaches revolving around rule based techniques have been dominating for this problem since long time. However, with the rise of deep learning based approaches, this perspective has changed as many deep neural models have outperformed conventional rule based approaches. In the  SIGMORPHON 2016 shared task \cite{cotterell-etal-2016-sigmorphon}of morphological (re)inflection, both neural and non-neural architecture based submissions were received. It was observed that the best-performed models were neural and they outperformed non-neural model by over 10\% in average accuracy.

Neural approaches shows good performance and they require very less feature engineering. Finite state machine based approaches require rule writing by human experts, which is expensive and time-consuming process \cite{klein-tsarfaty-2020-getting}. Statistical approaches such as SVM and CRF require heavy feature engineering while the neural approaches do not require any feature engineering. It has been shown by many authors that the neural approaches can be incorporated in different statistical machine learning methods \cite{rastogi2016weighting} \cite{vylomova2019contextualization}. From the results persented in various works, we observe that the recent approaches based on deep learning gives promising results given the availability of large training data. In case of unavailability of the large amount of data, often the rule based or statistical methods perform reasonable well.

For the low resource languages, building morphological analyzers can be a challenging task due to scarcity of the resources such as training data required for the deep neural network models \cite{bafna-zabokrtsky-2022-subword}.  Multilingual models have shown promising results for cross lingual transfer from high resource languages to low-resource languages \cite{wu-dredze-2019-beto} \cite{lauscher-etal-2020-zero} \cite{park-etal-2021-morphology} \cite{pawar-etal-2023-evaluating}. In \cite{kondratyuk-2019-cross}, authors suggest that if the multilingual BERT model is fine tuned on all available treebanks, then it can learn good cross lingual information which boost the accuracy of the lemmatization and morphological tagging tasks. Still, for few shot and zero shot learning, we require techniques that work with less amount of data and can address the limitations of multilingual models. One such method is to use multilingual models that are trained on large number of languages and are capable of using linguistic knowledge on the low resource language. Currently, one such approach is the use of continuous cross lingual representation space that encode the linguistic knowledge of multiple languages into a single space. Cross lingual transfer for lemmatization task in context of low resource Indian languages is studied in \cite{saunack-etal-2021-low}. The authors observe that even with very small number of training examples, the model gives decent results and the present of additional features such as POS tag can improve the accuracy of the system.
\begin{figure}
    \centering
    \includegraphics[scale=0.5]{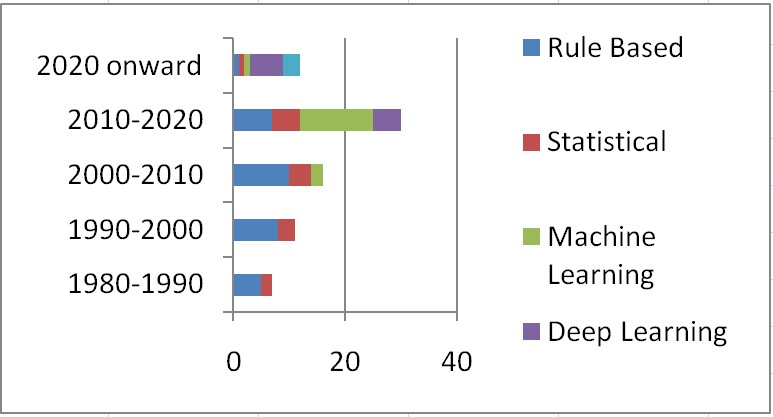}
    \caption{Decade wise popularity of various approaches for computational morphology related work (Based on number of publications)}
    \label{fig:decade_wise_approach}
\end{figure}
\begin{figure}
    \centering
    \includegraphics[scale=0.5]{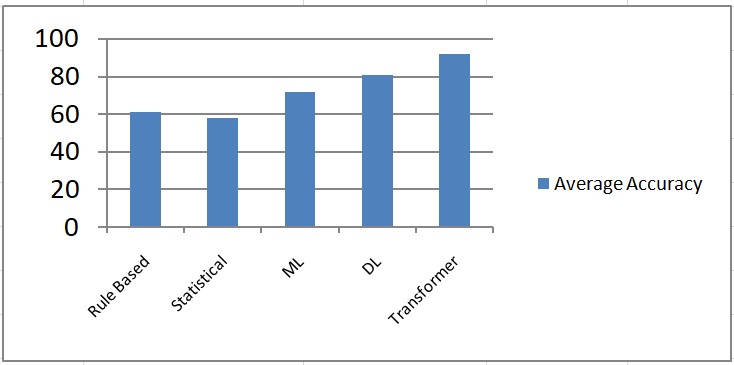}
    \caption{Average accuracies of various approaches for computational morphology related work }
    \label{fig:approach_wise_accuracy}
\end{figure}
\begin{figure}
    \centering
    \includegraphics[scale=0.5]{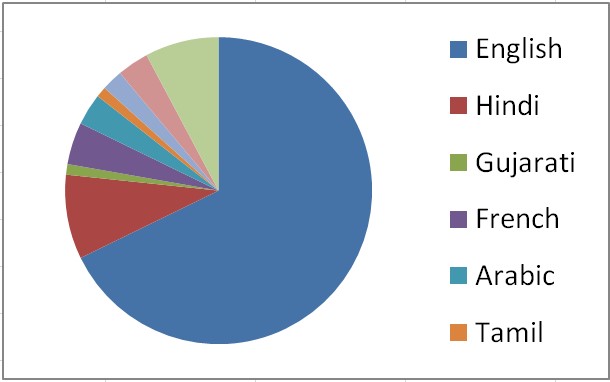}
    \caption{Percentage of work done for various languages related to computational morphology  }
    \label{fig:language_wise_paper}
\end{figure}
\section{Research Issues}
From the exhaustive survey of various techniques for computational morphology related problems, we identify below research issues :
\begin{itemize}
    \item In the recent times, neural models especially deep learning and transformer based approaches are dominating for the various tasks related to the computational morphology as highlighted in Figure \ref{fig:decade_wise_approach}. In general, neural models gives higher accuracy (Figure \ref{fig:approach_wise_accuracy}) but they are data-hungry and require large amount of labelled data for producing accurate results. This may not be possible for the low resource languages. Hence, more research should be done on how to utilize neural models with small training data.
    \item As evident from Figure \ref{fig:language_wise_paper}, most of the work related to morphology is focused on resource rich languages like English. For the low resource languages, transfer learning and data augmentation based models may be explored.
    \item Neural models do not require any feature engineering. However, they lack interpretability i.e what the model has learned. Research can be done on this aspect so that a better understanding can be developed about the contribution of various parts of the model in the overall output.
    \item The output of the neural model may be evaluated from the linguistic perspective to develop a better understanding about the learning of a model. Such knowledge helps in understanding how well the model captures the word formation process and may contribute to the overall performance of the system.
    \item Research contribution can be made towards making the datasets and resources publicly available for the low-resource language.
\end{itemize}
\section{Conclusion}
This survey provides exhaustive review of various approaches for the tasks related to the computational morphology. It provides an overview of different methods in chronological order. It explains the traditional rule based approaches such as FST, paradigm model etc in detail. Various supervised and unsupervised machine learning methods are also explored in this survey. The recent deep neural network based and transformer based models are also reviewed in this article. In the end, the discussion related to the effectiveness of deep neural models compared to the traditional approaches is carried out. We also investigate the tasks related to the computational morphology in the context of the low resource languages and survey about how the cross lingual transfer and multilingual models can boost the performance especially in the low resource settings. We conclude by highlighting some open research issues in the field. This survey article can be a good starting point for overall understanding of the computational morphology field and different approaches applied so far for solving various sub problems.

\end{document}